\documentclass[11pt,a4paper]{article}
\usepackage[hyperref]{acl2021}
\usepackage{times}
\usepackage{latexsym}

\usepackage{microtype}

\aclfinalcopy 

\usepackage{times}
\usepackage{latexsym}

\usepackage{danudefs}
\usepackage{algorithmic}
\usepackage{algorithm}
\usepackage{booktabs}
\usepackage{xspace}
\usepackage{url}
\usepackage{mathtools}
\usepackage{amsmath}
\usepackage{graphicx}
\usepackage{comment}
\usepackage{caption}
\usepackage{subcaption}
\usepackage{multirow}
\usepackage{enumerate}
\usepackage{esvect}
\usepackage{csvsimple}
\usepackage{microtype}
\usepackage{adjustbox}
\usepackage{multirow}
\usepackage{wrapfig}

\usepackage{array}
\newcolumntype{H}{>{\setbox0=\hbox\bgroup}c<{\egroup}@{}}

\usepackage[english]{babel}
\usepackage{hyperref}
\addto\captionsenglish{%
}
\addto\extrasenglish{%
}

\usepackage{xr}
\makeatletter
\newcommand*{\addFileDependency}[1]{
  \typeout{(#1)}
  \@addtofilelist{#1}
  \IfFileExists{#1}{}{\typeout{No file #1.}}
}
\makeatother
\newcommand*{\myexternaldocument}[1]{%
    \externaldocument{#1}%
    \addFileDependency{#1.tex}%
    \addFileDependency{#1.aux}%
}
\myexternaldocument{Supplementary}

\usepackage{todonotes}

\makeatletter
\if@todonotes@disabled

\else

\fi
\makeatother

\title{\emph{Unmasking the Mask} -- Evaluating Social Biases in \\Masked Language Models}

\author{
    Masahiro Kaneko\\
    Tokyo Institute of Technology \\
    {\tt masahiro.kaneko} \\
    {\tt @nlp.c.titech.ac.jp} \\
    \And
    Danushka Bollegala\Thanks{ Danushka Bollegala holds concurrent appointments as a Professor at University of Liverpool and as an Amazon Scholar. This paper describes work performed at the University of Liverpool and is not associated with Amazon.} \\
  University of Liverpool, Amazon\\
  {\tt danushka@liverpool.ac.uk}}

\date{}

\begin{document}
\maketitle
\begin{abstract}
Masked Language Models (MLMs) have shown superior performances in numerous downstream NLP tasks when used as text encoders.
Unfortunately, MLMs also demonstrate significantly worrying levels of social biases.
We show that the previously proposed evaluation metrics for quantifying the social biases in MLMs are problematic due to following reasons:
(1) prediction accuracy of the masked tokens itself tend to be low in some MLMs, which raises questions regarding the reliability of the evaluation metrics that use the (pseudo) likelihood of the predicted tokens, and 
(2) the correlation between the prediction accuracy of the mask and the performance in downstream NLP tasks is not taken into consideration, and 
(3) high frequency words in the training data are masked more often, introducing noise due to this selection bias in the test cases.
To overcome the above-mentioned disfluencies, we propose All Unmasked Likelihood (AUL), a bias evaluation measure that predicts  \emph{all} tokens in a test case given the MLM embedding of the \emph{unmasked} input. 
We find that AUL accurately detects different types of biases in MLMs.
We also propose AUL with Attention weights (AULA) to evaluate tokens based on their importance in a sentence.
However, unlike AUL and AULA, previously proposed bias evaluation measures for MLMs systematically overestimate the measured biases, and are heavily influenced by the unmasked tokens in the context.
\end{abstract}

\section{Introduction}
\label{sec:intro}

Masked Language Models~\cite[\textbf{MLM}s;][]{GPT-2,GPT3,BERT,RoBERTa} produce accurate text representations that can be used to obtain impressive performances in numerous downstream NLP applications as-is or by fine-tuning.
However, MLMs are also shown to encode worrying levels of social biases such as gender and racial biases~\cite{may-etal-2019-measuring,Zhao:2019a,Tan:2019}, which make it problematic when applied to tasks such as automatic summarisation or web search~\cite{Bender:2019}.
By detecting and quantifying the biases directly in the MLMs, we can address the problem at the source, rather than attempting to address it for every application that uses these pretarined MLMs.
Motivated by this need, we propose bias evaluation measures for MLMs.

We argue that an ideal bias evaluation measure for MLMs must satisfy the following two criteria.

\noindent\textbf{Criterion 1: The bias evaluation measure must consider the prediction accuracy of the MLM under evaluation.}

For example, if the MLM has low accuracy when predicting a masked token in a sentence, then using its pseudo-likelihood as an evaluation measure of bias is unreliable when distinguishing between stereotypical vs. anti-stereotypical sentences~\cite{Nadeem:2020,crows-pairs}.
MLMs can often predict multiple plausible tokens for a given context (e.g. \emph{The chess player was} [MASK].), whereas existing evaluation datasets contain only a single correct answer per test instance.
Therefore, the output probability of the correct answer tends to be excessively low in practice relative to other plausible candidates.
Consequently, as we later show in \autoref{sec:token-prediction}, the performance of pseudo-likelihood-based bias evaluation measures significantly deteriorate when there exist multiple valid answers to a given test instance.


\noindent\textbf{Criterion 2: When we apply a particular mask and predict a token, we must consider any biases introduced by the other (unmasked) words in the context}.

For computational tractability, previously proposed  pseudo-likelihood-based scoring methods~\cite{Nadeem:2020,crows-pairs} assumed that the masked tokens are statistically independent. 
However, this assumption does not hold in reality and introduces significant levels of noises to the evaluation measures.
It is noteworthy that not all downstream tasks that use MLMs use masks for predicting tokens.
For example, downstream tasks that use MLMs for representing input texts such as a sentence-level sentiment classifier~\cite{BERT} would use the sentence embeddings obtained from an MLM instead of using it to predict the input tokens.
Therefore, we argue that it is undesirable for any biases associated with the masked tokens to influence the bias evaluation of an MLM.
Ideally, we must distinguish between the intrinsic biases embedded in an MLM vs. the biases that creep in during task-specific fine-tuning.
The focus of this paper is evaluating the former intrinsic biases in MLMs.

We propose \textbf{All Unmasked Likelihood} (AUL)\footnote{The code is publicly available at \url{https://github.com/kanekomasahiro/evaluate_bias_in_mlm}}, a bias evaluation measure that predicts  \emph{all} of the tokens in a test sentence given the MLM embedding of its \emph{unmasked} input.
AUL satisfies both criteria and overcomes the disfluencies in the prior MLM bias evaluation measures.
First, using the MLM under evaluation, we create an embedding for a test sentence \emph{without} masking any of its tokens, thereby using information related to all of the tokens in that sentence.
Second, by requiring the MLM to simultaneously predict \emph{all} of the unmasked tokens in a sentence, we avoid any selectional biases due to masking a subset of the input tokens, such as highly frequent words.

AUL evaluates biases by considering all tokens equally, however, each token in a sentence has different importance. 
For example, tokens such as articles and prepositions have less importance.
It is not desirable for the likelihood of such tokens to affect the bias evaluation.
Therefore, we propose \textbf{AUL with Attention weights} (AULA), which evaluates the bias by considering the weight of MLM attention as the importance of tokens.

We compare AUL and AULA against previously proposed MLM bias evaluation measures by~\newcite{Nadeem:2020} on the StereoSet (\textbf{SS}) dataset  and by~\newcite{crows-pairs} on the CrowS-Pairs (\textbf{CP}) dataset.
Experimental results show that AUL\footnote{AULA has the similar accuracy scores as AUL} outperform prior proposals, reporting higher accuracies for predicting the tokens in test sentences (\autoref{sec:token-prediction}).
This is particularly critical for SS where there is only one designated correct answer per test sentence, reporting 95.71 points drop in accuracy compared to AUL.
Moreover, we show that the token prediction accuracy under AUL is sensitive to the meaningful associations in the input sentence by randomly shuffling the tokens in sentence or by replacing a word with an unrelated one (\autoref{sec:meaningful}).
This result shows that AUL can distinguish between natural sentences in a language from meaningless ones.
This is a desirable  property because it shows that AUL is sensitive to the language modelling ability of the MLM.
As we later see in \autoref{sec:adv-freq}, words in the \emph{advantaged groups}~\cite{crows-pairs} tend to occur in a corpus statistically significantly more than the words in \emph{disadvantaged groups}.
This adversely affects previously proposed evaluation measures, rendering their bias evaluations less reliable compared to AUL and AULA.

We evaluated the performance by comparing the existing methods and the proposed methods with the human bias score.
As a result, it was shown that AUL and AULA outperform the existing evaluation methods in \autoref{sec:bias-eval}.
Especially it is true for AULA.
Although we still find unfair biases in MLMs according to AUL and AULA, we note that these levels are less than what had been reported in prior work~\cite{kurita-etal-2019-measuring,Nadeem:2020,crows-pairs}.

\section{Related Work}
\label{sec:related}

Our focus in this paper is evaluating and \emph{not} proposing methods to mitigate the biases in MLMs.
Therefore, we primarily discuss prior work on evaluation metrics and benchmarks for social biases.
For details on debiasing methods for MLMs see~\cite{Kaneko:EACL:2021b,schick2020self,liang-etal-2020-monolingual}.

\subsection{Biases in Static Embeddings}
\newcite{Tolga:NIPS:2016} use word analogies to evaluate gender bias in pretrained static word embeddings~\cite{Glove,Milkov:2013}. 
If an embedding predicts stereotypical analogies such as \emph{doctor} : \emph{man} :: \emph{woman} : \emph{nurse}, they conclude the embedding to be gender-biased.
\newcite{manzini-etal-2019-black} extend this method to other stereotypical biases such as racial and religious biases.

The Word Embedding Association Test~\cite[\textbf{WEAT};][]{WEAT} imitates the human Implicit Association Test~\cite[\textbf{IAT};][]{IAT} for word embeddings, where the association between two sets of target concepts (e.g. European American vs. African American names) and two sets of attributes (e.g.  Pleasant (\emph{love, cheer, peace}) vs.  Unpleasant (\emph{ugly, evil, murder}) attributes).
Here, the association is measured using the cosine similarity between word embeddings.
\newcite{ethayarajh-etal-2019-understanding} showed that WEAT systematically overestimates biases and proposed relational inner product association (RIPA), a subspace projection method, to overcome this problem.

Word Association Test~\cite[\textbf{WAT};][]{du-etal-2019-exploring} measures gender bias over a large set of words.
WAT calculates a gender information vector for each word in an association graph~\cite{Deyne2019TheW} by propagating~\cite{Zhou2003LearningWL} information related to masculine and feminine words.
SemBias dataset~\cite{Zhao:2018ab} contains gender-definitional,  gender-stereotypical and gender-unrelated word-pairs. 
The ability to resolve gender-related pronouns without unfair biases has been used as an evaluation measure.
 WinoBias~\cite{Zhao:2018aa} and OntoNotes~\cite{weischedel2013ontonotes} datasets are used for evaluating the social biases of word embeddings under coreference resolution.

\subsection{Biases in Contextualised Embeddings}
Social biases have been identified not only in static word embeddings but also in contextualised word embeddings produced by MLMs~\cite{bommasani-etal-2020-interpreting,karve-etal-2019-conceptor,Dev:2019}.
Sentence Encoder Association Test~\cite[\textbf{SEAT};][]{may-etal-2019-measuring} extends WEAT to sentence encoders by creating artificial sentences using templates such as ``\emph{This is [target]}'' and ``\emph{They are [attribute]}''. 
Next, different sentence encoders are used to create embeddings for these artificial sentences, and cosine similarity between the sentence embeddings is used as the association metric.
However, they did not find any clear indication of biases for ELMo~\cite{Elmo} and BERT~\cite{BERT}.
\newcite{kurita-etal-2019-measuring} showed that cosine similarity is not suitable as an evaluation measure for SEAT and proposed the log-odds of the target and prior probabilities of the sentences computed by masking  respectively only the target vs. both target and attribute.
In addition, occupation templates~\cite[\textbf{OCCTMP};][]{liang-etal-2020-monolingual} evaluate the gender bias of MLMs by comparing the difference in the log-likelihood between \emph{he} and \emph{she} in the masked token of the template ``[MASK] \emph{is a/an [occupation]}'' created from occupation words (e.g. \emph{doctor, nurse, dancer}).

Using artificial contexts~\cite{liang-etal-2020-monolingual,may-etal-2019-measuring,kurita-etal-2019-measuring} for evaluating biases in MLMs have several drawbacks such as
(a) artificial contexts not reflecting the natural usage of a word,
(b) requiring the stereotypical attribute terms to be predefined, 
and (c) being limited to single word target terms. 
To address these drawbacks \newcite{Nadeem:2020} crowdsourced, StereoSet (SS), a dataset for associative contexts covering four types of stereotypical biases: race, gender, religion, and profession. 
SS contains test instance both at intrasentence and intersentence discourse levels.
They proposed a Context Association Test (CAT) for evaluating both language modelling ability as well as the stereotypical biases of pretrained MLMs.
In CAT, given a context containing a target group (e.g. \emph{housekeeper}), they provide three different ways to instantiate its context corresponding to a stereotypical, anti-stereotypical or an unrelated association.

\newcite{crows-pairs} created Crowdsourced Stereotype Pairs benchmark (CP) covering nine types of social biases. 
Test instances in CP consist of sentence pairs where one sentence is more stereotypical than the other.
Annotators are instructed to write examples that demonstrate stereotypes contrasting historically disadvantaged groups against advantaged groups.
They found that the test instances in CP to be more reliable than the ones in SS via a crowdsourced validation task.
In CP, the likelihood of the unmodified tokens between the two sentences in a test sentence-pair, given their modified tokens, is used to estimate the preference of an MLM to select a stereotypical sentence over a less stereotypical one.
This is in contrast to SS, where the likelihood of the modified tokens given the unmodified tokens was used to determine the preference of an MLM.
However, masking tokens from the test sentences and predicting only those masked tokens (as opposed to all tokens in the sentence) prevents the MLM from producing accurate sentence embeddings and favours advantaged groups, which tend to be more frequent than the disadvantaged groups in text corpora used to train MLMs.
On the other hand, AUL overcomes those limitations in the previous bias evaluation measures for MLMs by not masking any tokens from a test sentence and predicting all tokens (as opposed to a subset of masked tokens) in the sentence.



\section{All Unmasked Likeihood}
\label{sec:AUL}



Let us consider a test sentence $S = w_0, w_1, \ldots, w_{|S|}$, containing length $|S|$ sequence of tokens $w_i$, where part of $S$ is modified to create a stereotypical (or lack of thereof) example for a particular social bias. 
For example, consider the sentence-pair ``\emph{\textbf{John} completed \textbf{his} PhD in machine learning}''  vs. ``\emph{\textbf{Mary} completed \textbf{her} PhD in machine learning}''.
The modified tokens for the first sentence are \{\emph{John}, \emph{his}\}, whereas for the second sentence they are \{\emph{Mary}, \emph{her}\}.
On the other hand, the unmodified tokens between two sentences are \{\emph{completed}, \emph{PhD}, \emph{in}, \emph{machine}, \emph{learning}\}.

For a given sentence $S$, let us denote its list of modified tokens by $M$ and unmodified tokens by $U$ such that $S = M \cup U$ is the list of all tokens in $S$.\footnote{Note that we consider lists instead of sets to account for multiple occurrences of the same word in a sentence.}
In SS, $M$ and $U$ are specified for each test sentence, whereas in CP they are determined given a test sentence-pair.

Given an MLM with pretrained parameters $\theta$, which we must evaluate for its social biases, let us denote the probability $P_{\mathrm{MLM}}(w_i | S_{\setminus w_{i}}; \theta)$ assigned by the MLM to a token $w_i$ conditioned on the remainder of the tokens, $S_{\setminus w_{i}}$. 
Similar to using log-probabilities for evaluating the naturalness of sentences using conventional language models, \newcite{salazar-etal-2020-masked} showed that, $\mathrm{PLL}(S)$, the pseudo-log-likelihood (PLL) score of sentence $S$ given by \eqref{eq:PLL}, can be used for evaluating the preference expressed by an MLM for $S$.
\begin{align}
    \label{eq:PLL}
    \mathrm{PLL}(S) \coloneqq \sum_{i=1}^{|S|} \log P_{\mathrm{MLM}}(w_i | S_{\setminus w_{i}}; \theta)
\end{align}
PLL scores can be computed out of the box for MLMs and are more uniform across sentence lengths (no left-to-right bias), which enable us to recognise natural sentences in a language~\cite{wang-cho-2019-bert}.
PLL can be used in several ways to define bias evaluation scores for MLMs as we discuss next.

\newcite{Nadeem:2020} used, $P(M | U; \theta)$, the probability of generating the modified tokens given the unmodified tokens in $S$.
We name this StereoSet Score (\textbf{SSS}) and is given by \eqref{eq:SSS}.
\begin{align}
    \label{eq:SSS}
    \mathrm{SSS}(S) \coloneqq  \frac{1}{|M|} \sum_{w \in M} \log P_{\mathrm{MLM}}(w | U; \theta)
\end{align}
Here, $|M|$ is length of $M$.
However, SSS is problematic because when comparing $P(M | U; \theta)$ for modified words such as \emph{John}, we could have high probabilities simply because such words have high frequency of occurrence in the data used to train the MLM and not because the MLM has learnt a social bias.

To address this frequency-bias in SSS, \newcite{crows-pairs} used $P(U|M; \theta)$ to define a scoring formula given by \eqref{eq:CPS}, which we refer to as the CrowS-Pairs Score (\textbf{CPS}).
\begin{align}
    \label{eq:CPS}
     \mathrm{CPS}(S) \coloneqq \sum_{w \in U} \log P_{\mathrm{MLM}}(w | U_{\setminus w}, M; \theta)
\end{align}
However, when we mask one token $w$ at a time from $U$ and predict it, we are effectively changing the context $(U_{\setminus w}, M)$ used by the MLM as the input.
This has two drawbacks. 
First, the removal of $w$ from the sentence results in a loss of information that the MLM can use for predicting $w$.
Therefore, the prediction accuracy of $w$ can decrease, rendering the bias evaluations unreliable.
This violates Criterion 1 in \autoref{sec:intro}.
Second, even if we remove one token $w$ at a time from $U$, the remainder of the tokens $\{U_{\setminus w}, M\}$ can still be biased.
Moreover, the context on which we condition the probabilities continuously vary across predictions.
This violates Criterion 2 in \autoref{sec:intro}.

To overcome the above-mentioned disfluencies in previously proposed MLM bias evaluation measures, we propose a simple two-step solution.
First, instead of masking out tokens from $S$, we provide the complete sentence to the MLM.
Second, we predict all tokens in $S$ that appear between begin and end of sentence tokens.
Specifically, we apply Byte Pair Encoding~\cite[BPE;][]{Sennrich:2016} to $S$ to (sub)tokenise it, and require the MLM to predict exactly the same number of (sub)tokens as we have in $S$ during the prediction step.
We name our proposed evaluation measure as All Unmasked Likelihood (\textbf{AUL}) and calculate using \eqref{eq:AUL}.
\begin{align}
    \label{eq:AUL}
    \mathrm{AUL}(S) \coloneqq \frac{1}{|S|} \sum_{i=1}^{|S|} \log P_{\mathrm{MLM}}(w_i | S; \theta)
\end{align}
At a first glance one might think that we can predict $w_i$ with absolute confidence (i.e. $\forall_{w_{i}}, \ P_{\mathrm{MLM}}(w_i | S; \theta) = 1$) because $w_i \in S$. However, in MLMs this is not the case because some lossy compressed representation (e.g. an embedding of $S$) is used during the prediction of $w$.

Moreover, we calculate the likelihood considering the attention weights to evaluate bias considering importance of words in a sentence.
AUL with Attention weights (\textbf{AULA}) is calculated as follows:
\begin{align}
    \label{eq:AULA}
    \mathrm{AULA}(S) \coloneqq \frac{1}{|S|} \sum_{i=1}^{|S|} \alpha_i \log P_{\mathrm{MLM}}(w_i | S; \theta)
\end{align}
Here, $\alpha_i$ is the average of all multi-head attentions associated with $w_i$.

Given a score function $f \in \{\mathrm{SSS}, \mathrm{CPS}, \mathrm{AUL}, \mathrm{AULA}\}$, we use the percentage of stereotypical ($S^{st}$) test sentences preferred by the MLM over anti-stereotypical ($S^{at}$) ones to define the corresponding bias evaluation measure (\textbf{bias score}) as follows:
\begin{align}
    \label{eq:score}
  \frac{1}{N}\sum_{(S^{st}, S^{at})} \mathbb{I}(f(S^{st}) > f(S^{at}))
\end{align}
Here, $\mathbb{I}$ is the indicator function, which returns $1$ if its argument is True and $0$ otherwise, and $N$ is the total number of test instances.
According to this evaluation measure values close to 50 indicate that the MLM under evaluation is neither stereotypically nor anti-stereotypically biased, hence, can be regarded as unbiased. 
On the other hand, values below 50 indicate a bias towards the anti-stereotypical group, whereas values above 50 indicate a bias towards the stereotypical group.

\section{Experiments and Findings}

\subsection{Experimental Setup}

In our experiments, we use the following MLMs: BERT ~\cite[\textbf{bert-base-cased};][]{BERT}, RoBERTa~\cite[\textbf{roberta-large};][]{RoBERTa} and ALBERT~\cite[\textbf{albert-large-v2};][]{ALBERT}\footnote{The parameter settings are given in \url{https://huggingface.co/transformers/pretrained_models.html}}.
We used the MLM implementations in the transformer library~\cite{wolf-etal-2020-transformers}\footnote{\url{https://github.com/huggingface/transformers}}.

We used the publicly available CP dataset\footnote{\url{https://github.com/nyu-mll/crows-pairs}}, which is crowdsourced and annotated by workers in the United States.
This dataset contains 1,508 sentence-pairs covering nine social bias types: \emph{race} (516), \emph{gender} (262), \emph{sexual orientation} (84), \emph{religion} (105), \emph{age} (87), \emph{nationality} (159), \emph{disability} (60), \emph{physical appearance} (63), and \emph{socioeconomic status} (172). Here, we show the number of sentence-pair instances in brackets.
Each sentence-pair is further classified depending on whether it is biased towards the advantaged group (e.g. \textit{\textbf{She/He} addressed the shareholders as the CEO of the company.}), or the disadvantaged group (e.g. \textit{\textbf{Women/Men} are always too sensitive about things}).

Because the test portion of the SS dataset is publicly unavailable, we used its development set\footnote{\url{https://github.com/moinnadeem/StereoSet}}.
In addition to the association tests that predict masked tokens for measuring bias at sentence level (\textbf{Intrasentence}), SS also has association tests that evaluate the social biases by predicting an appropriate context sentence at discourse level (\textbf{Intersentence}).
However, in our experiments, we use only Intrasentence association tests from SS, and do not use Intersentence association tests because this set does not use masks for bias evaluation.
SS dataset contains 2,106 sentence-pairs covering four types of social biases: \emph{gender} (255), \emph{profession} (810), \emph{race} (962) and  \emph{religion} (79).
Moreover, \underline{unrelated words} (e.g. \textit{The chess player was \underline{fox}}.) are also used as candidates to evaluate the validity of an MLM's predictions.
Unlike in CP, in SS sentences are not classified into advantaged vs. disadvantaged groups.

We use CPS (\autoref{eq:CPS}) as the scoring formula  with the CP dataset, whereas SSS (\autoref{eq:SSS}) is used with the the SS dataset.
The proposed evaluation measures, AUL and AULA, can be used with both CP and SS datasets to separately compute MLM bias scores, denoted respectively by \textbf{AUL (CP)}, \textbf{AUL (SS)}, \textbf{AULA (CP)} and \textbf{AULA (SS)}.
All experiments were conducted on a GeForce GTX 1080 Ti GPU.
Evaluations on both CP and SS are completed within fifteen minutes.

\subsection{Token Prediction Accuracy}
\label{sec:token-prediction}

\begin{table}[t!]
\centering
\scalebox{0.75}{
\begin{tabular}{lcccc}
\toprule
MLM & CPS & AUL (CP) & SSS & AUL (SS) \\
\midrule
BERT & 62.98 & \textbf{82.76}$^{\dagger}$ & 2.20 & \textbf{92.16}$^{\dagger}$ \\
RoBERTa & 68.11 & \textbf{99.54}$^{\dagger}$ & 3.17 & \textbf{98.88}$^{\dagger}$ \\
ALBERT & 56.20 & \textbf{88.01}$^{\dagger}$ & 2.21 & \textbf{81.19}$^{\dagger}$ \\
\bottomrule
\end{tabular}
}
\caption{Token prediction accuracy of previously proposed MLM bias evaluation measures (CPS, SSS) and the proposed AUL measure on CP and SS datasets. $\dagger$ indicates statistically significant scores according to the McNemar's test ($p < 0.01$).}
\label{tbl:mask_acc}
\vspace{-3mm}
\end{table}

First, we show that the prediction accuracy of a masked token under the previously proposed MLM bias evaluation measures (e.g. CPS, SSS) is lower than that of the proposed evaluation measures, AUL and AULA. 
 Note that multiplying the attention weights by the likelihood does not affect the token prediction accuracy within a sentence, hence AUL and AULA have the same token prediction accuracy.
 Therefore, both AUL and AULA are denoted as AUL for in the experimental results reported in this section related to token prediction accuracy.

Typically MLMs are trained using subtokenised texts and the subtokenisation of a word is not unique.
In CP, we measure the prediction accuracy of the unmodified tokens between the two sentences in a sentence-pair.
Therefore, the number of subtokens to be predicted is the same between the two sentences in a sentence-pair in CP.
However, for the intrasentence test cases in SS, we must select between a stereotypical and an anti-stereotypical candidate to fill the masked slot in a sentence, while the remaining context in the sentence is held fixed.
If the number of subtokens is the same for both candidates in a test sentence, we consider the prediction to be accurate, if the predicted sequence of subtokens exactly matches at least one of the two candidates (i.e. stereotype and anti-stereotype).
However, if the number of subtokens in each candidate is different, we insert masked slots matching the number of subtokens in each candidate and predict all those slots.

\begin{table*}[t!]
\footnotesize
\centering
\begin{adjustbox}{width=\linewidth}
\begin{tabular}{lcccccccccc}
\toprule
            & Adv & Dis & 1 & 2 & 3 & 4 & 5 & 6 & 7 & 8 \\
\midrule
 Race & \textbf{3.75} & 5.25 & \underline{american} & \underline{james} & \textit{african} & \textit{asian} & \underline{carl} & \textit{tyrone} & \underline{caucasian} & \textit{jamal} \\
 Gender & \textbf{3.75} & 5.25 & \underline{he} & \underline{his} & \textit{her} & \textit{she} & \underline{men} & \textit{woman} & \underline{him} & \textit{women} \\
 Sexual orientation & \textbf{3.5} & 5.5 & \underline{woman} & \underline{wife} & \underline{husband} & \textit{gay} & \textit{lesbian} & \textit{homosexual} & \textit{bisexual} & \underline{heterosexual} \\
 Religion & \textbf{4.25} & 4.75 & \underline{church} & \underline{christian} & \textit{jewish} & \textit{muslim} & \textit{muslims} & \underline{christians} & \textit{jew} & \underline{atheist} \\
 Age & \textbf{4} & 5 & \textit{old} & \underline{young} & \underline{middle} & \textit{boy} & \underline{aged} & \underline{adults} & \textit{elderly} & \textit{teenagers} \\
 Nationality & \textbf{3} & 6 & \underline{american} & \underline{canada} & \underline{canadian} & \textit{chinese} & \textit{italian} & \underline{americans} & \textit{mexican} & \textit{immigrants} \\
 Disability & \textbf{3} & 6 & \underline{normal} & \underline{smart} & \underline{healthy} & \textit{ill} & \textit{mentally} & \underline{gifted} & \textit{autistic} & \textit{retarded} \\
 Physical appearance & \textbf{4} & 5 & \textit{short} & \underline{beautiful} & \underline{tall} & \underline{thin} & \textit{ugly} & \textit{fat} & \underline{skinny} & \textit{overweight} \\
 Socioeconomic status & \textbf{4} & 5 & \textit{poor} & \underline{doctor} & \underline{rich} & \textit{poverty} & \underline{wealthy} & \underline{businessman} & \textit{homeless} & \textit{ghetto} \\
\bottomrule
\end{tabular}
\end{adjustbox}
\caption{The mean rank of each group and the descending order of each word by the frequency of occurrence in Wikipedia \& BookCorpus with four high-frequency words in the the advantaged group (Adv) and disadvantaged group (Dis) group in CP. The \underline{underline} represents the words that belong to the advantaged group, and the \textit{italics} represent the words that belong to the disadvantaged group.}
\label{tbl:word_freq}
\end{table*}

\begin{table*}[t]
\centering
\scalebox{0.75}{
\begin{tabular}{lccccccccc}
\toprule
     & \multicolumn{3}{c}{All Masked (CP)} & \multicolumn{3}{c}{AUL (CP)} & \multicolumn{3}{c}{AULA (CP)} \\
     \cmidrule(lr){2-4} \cmidrule(lr){5-7} \cmidrule(lr){8-10}
 MLM & Adv & Dis & $|$Diff$|$ & Adv & Dis & $|$Diff$|$ & Adv & Dis & $|$Diff$|$ \\
\midrule
BERT    & 54.13 & 47.36 & 6.77  & 49.54 & 53.49 & \textbf{3.95}  & 50.46 & 54.65 & 4.19 \\
RoBERTa & 65.14 & 37.05 & 28.09 & 51.38 & 64.26 & 12.88 & 51.83 & 60.78 & \textbf{8.95} \\
ALBERT  & 55.05 & 45.35 & 9.70  & 55.05 & 52.95 & 2.10  & 54.13 & 52.87 & \textbf{1.26} \\
\bottomrule
\end{tabular}
}
\caption{Bias score for the advantaged group (Adv) and disadvantaged group (Dis) in CP when all tokens are masked (All Masked (CP)) and when all tokens are not masked (AUL (CP) and AULA (CP)). $|\textrm{Diff}|$ is the absolute value of the difference between Adv and Dis.}
\label{tbl:mask_bias}
\end{table*}

For example, consider the SS instance ``\textit{The chess player was \underline{\hspace{0.5cm}}'' \{stereotypical=\textbf{hispanic}, anti-stereotypical=\textbf{asian}\}},
where \textit{hispanic} is split into the subtokens \textit{his}+\textit{panic}, and \textit{asian} into \textit{asi}+\textit{an}.
In this case, where the number of subtokens is the same for both candidates, both candidates can be predicted from the same masked input:  ``\textit{The chess player was} [MASK] [MASK]''.
On the other hand, let us assume that \textit{hispanic} is split into the three subtokens \textit{his}+\textit{pa}+\textit{nic}, and \emph{asian} into the two subtokens \textit{asi}+\textit{an}.
In this case, because the numbers of subtokens are different for the two candidates, we use the input ``\textit{The chess player was} [MASK] [MASK] [MASK]'' for predicting the stereotypical candidate  and ``\textit{The chess player was} [MASK] [MASK]'' for predicting the anti-stereotypical candidate.
Among the 2,106 Intrasentence test cases in SS, the numbers of instances with equal number of subtokens for the two candidates are 1,298, 1,509 and 1,490 respectively under the subtokenisers used in BERT, RoBERTa, and ALBERT.

\autoref{tbl:mask_acc} shows the token prediction accuracies in CP (CPS and AUL (CP)) and SS (SSS and AUL (SS)) datasets.
For all MLMs compared, we see that AUL significantly outperforms the previously proposed CPS and SSS measures.
Interestingly, the token prediction accuracy of SSS, which targets different modified tokens with the same context, is particularly low.
This shows that AUL is robust even in the presence of multiple plausible candidates.
Therefore, Criterion 1 is better satisfied by AUL compared to CPS and SSS. 
Note that the prediction accuracy of AUL given unmasked tokens as the input is not 100\%.
This suggests that the MLMs are trained to discard information from the input tokens.
The lower prediction accuracies of BERT and ALBERT compared to RoBERTa indicate that this loss of information is more prominent for those models.

\subsection{Word Frequency and Social Biases}
\label{sec:adv-freq}

Frequency of a word has shown to directly influence the semantic representations learnt for that word~\cite{Arora:TACL:2016,Schick:2020}.
To understand how word frequency influences PLL-based bias evaluation measures, we examine the frequency of words in the advantaged and disadvantaged groups on a corpus that contains Wikipedia articles\footnote{Wikipedia dump on 2018 Sept is used.} \& BookCorpus~\cite{Zhu_2015_ICCV}, popularly used as MLM training data.
This corpus contains a total of 3 billion tokens.
For each bias type in CP, we find the frequency of the words in the corresponding advantaged and disadvantaged groups in this corpus.\footnote{SS does not split test instances into advantaged vs. disadvantaged groups, hence excluded from this experiment.}
Words that have non-stereotypical senses (e.g. \emph{white} and \emph{black} are used as colours) are ignored from this analysis.
For words that appear in both groups, we assign them to the group with the higher frequency.

\autoref{tbl:word_freq} shows the mean rank of the words that belong to each group for different social bias categories in CP.
Moreover, we show the top 8 frequent words across advantaged (underlined) and disadvantaged groups.\footnote{See Supplementary for the raw frequency counts.}
From \autoref{tbl:word_freq}, we see that the mean rank for the advantaged group is higher than that for the disadvantaged group in all bias categories.
This shows that compared to the words in the disadvantaged groups, words in the advantaged group have a higher frequency of occurrences in the corpora used to train MLMs.

\begin{table}[t]
\centering
\scalebox{0.75}{
\begin{tabular}{lccc}
\toprule
 & AUL (CP)  & \multicolumn{2}{c}{AUL (SS)} \\
 \cmidrule(lr){2-2} \cmidrule(lr){3-4}
MLM & Shuffled & Shuffled & Unrelated \\
\midrule
BERT & 69.63$^{\dagger}$ (-13.13) & 62.30$^{\dagger}$ (-29.86) & 71.67 (-20.49) \\
RoBERTa & 80.82$^{\dagger}$ (-18.72) & 76.49$^{\dagger}$ (-22.39) & 93.88 (-5.00) \\
ALBERT & 80.86$^{\dagger}$ (-7.15) & 73.18$^{\dagger}$ (-8.01) & 76.08 (-5.11) \\
\bottomrule
\end{tabular}
}
\caption{Token-level prediction accuracy of MLMs for randomly shuffled (in CP and SS) and unrelated (in SS) sentences are shown for AUL. Relative drop in accuracy w.r.t. when using the original sentence (reported in \autoref{tbl:mask_acc}) is shown in brackets.
 $\dagger$ denotes significance drops according to the McNemar's test ($p < 0.01$). 
 For Unrelated, the number of subtokens with unrelated word may be different from the original sentence, thus significant difference tests cannot be performed.}
\label{tbl:mask_noise}
\end{table}

\begin{table*}[t]
\centering
\scalebox{0.75}{
\begin{tabular}{lcccccc}
\toprule
MLM & CPS & AUL (CP) & AULA (CP) & SSS & AUL (SS) & AULA (SS) \\
\midrule
BERT & 58.62 & 52.92 & 54.05 & 57.26 & 50.28 & 51.38 \\
RoBERTa & 65.45 & 62.40 & 59.48  & 61.97 & 59.07 & 55.98 \\
ALBERT & 60.41 & 53.25 & 53.05 & 58.88 & 58.07 & 58.31 \\
\bottomrule
\end{tabular}
}
\caption{Bias scores reported by CPS, SSS and AUL on CP and SS datasets for BERT, RoBERTa and ALBERT.}
\label{tbl:bias_eval}
\end{table*}

\begin{table}[ht!]
\centering
\scalebox{0.65}{
\begin{tabular}{llccc}
\toprule
MLM & Bias type & CPS & AUL (CP) & AULA (CP) \\
\midrule
\multirow{9}{*}{BERT} & Race & 54.26 & 52.33 & 55.23 \\
& Gender & 57.63  & 53.05  & 53.82 \\
& Sexual orientation & 65.48 & 58.33  & 57.14 \\
& Religion & 66.67 & 57.14  & 53.33 \\
& Age & 58.62 & 47.13  & 45.98 \\
& Nationality & 59.75 & 54.72 & 56.60 \\
& Disability & 68.33 & 73.33 & 75.00 \\
& Physical appearance & 63.49 & 57.14  & 50.79 \\
& Socioeconomic status & 58.72 & 41.86  & 45.35 \\
\midrule
\multirow{9}{*}{RoBERTa} & Race & 64.15 & 62.21 & 59.50 \\
& Gender & 58.40 & 58.02 & 59.54 \\
& Sexual orientation & 64.29 & 63.10 & 50.00 \\
& Religion &74.29  & 78.10 & 68.57 \\
& Age & 71.26 & 62.07 & 60.92 \\
& Nationality & 66.67 & 58.49 & 61.64 \\
& Disability & 70.00 & 66.67 & 61.67 \\
& Physical appearance & 73.02 & 66.67 & 63.49 \\
& Socioeconomic status & 66.86 & 60.47 & 53.49 \\
\midrule
\multirow{9}{*}{ALBERT} & Race & 59.11 & 50.19 & 49.81 \\
& Gender & 56.11 & 48.47 & 46.95 \\
& Sexual orientation & 71.43 & 69.05 & 69.05 \\
& Religion & 76.19 & 56.19 & 56.19 \\
& Age & 55.17 & 39.08 & 42.53 \\
& Nationality & 61.64 & 55.97 & 55.35 \\
& Disability & 73.33 & 68.33 & 63.33 \\
& Physical appearance & 61.90 & 66.67 & 68.25 \\
& Socioeconomic status & 52.33 & 54.65 & 56.40 \\
\bottomrule
\end{tabular}
}
\caption{Bias scores of CPS and AUL for the different types of biases in the CP dataset.}
\label{tbl:bias_cp}
\end{table}

\begin{table}[t]
\centering
\scalebox{0.75}{
\begin{tabular}{llccc}
\toprule
MLM & Bias type & SSS & AUL (SS) & AULA (SS) \\
\midrule
\multirow{4}{*}{BERT} & Gender & 63.14 & 49.80 & 48.63 \\
& Profession & 60.00 & 49.26 & 48.40 \\
& Race & 53.33 & 51.56 & 54.57 \\
& Religion & 58.23 & 46.84 & 51.90 \\
\midrule
\multirow{4}{*}{RoBERTa} & Gender & 73.33 & 62.75 & 54.12 \\
& Profession & 63.21 & 61.11 & 57.41 \\
& Race & 58.21 & 56.65 & 55.41 \\
& Religion & 58.23 & 55.70 & 54.43 \\
\midrule
\multirow{4}{*}{ALBERT} & Gender & 63.53 & 64.31 & 62.35 \\
& Profession &  60.49 & 60.00 & 61.11 \\
& Race & 56.34 & 55.09 & 54.99 \\
& Religion & 58.23 & 54.43 & 56.96 \\
\bottomrule
\end{tabular}
}
\caption{Bias scores of SSS and AUL for the different types of biases in the SS dataset.}
\label{tbl:bias_ss}
\end{table}

Recall that AUL and AULA do not mask any tokens in a test sentence, whereas CPS masks unmodified tokens one at a time and use the remaining tokens in the sentence to predicted the masked out token.
According to Criteria 2, an ideal MLM bias evaluation measure must not be influenced by the biases in the masked tokens.
To study the influence of the word frequency distribution of the masked tokens on MLM bias evaluation measures, we compare \textbf{AUL (CP)} and \textbf{AULA (CP)} (which do not mask input tokens) against \textbf{All Masked (CP)} baseline, where we mask all tokens from the sentence and predict those masked tokens on the CP dataset. 
If the masked tokens are biased, the score will be biased even though all tokens are masked.

From \autoref{tbl:mask_bias} we see that compared to AUL (CP) and AULA (CP), All Masked (CP) tends to overestimate the biases in the advantaged group, while underestimating the biases in the disadvantaged group.
As discussed in \autoref{tbl:word_freq}, the relatively high frequency of the advantaged group results in high bias scores under CPS, leading to an overestimate of social biases, whereas the reverse is true for the disadvantaged group.
Underestimating the social biases in disadvantaged groups by CPS is particularly worrying considering the fact that people belonging to the disadvantaged groups are already facing adverse consequences due to social biases.
On the other hand, we see that AUL (CP) and AULA (CP) consistently report biases in both groups.
Moreover, the absolute difference between the bias scores for the advantaged and disadvantaged groups (shown by $|\textrm{Diff}|$) is relatively small for AUL (CP) and AULA (CP) than All Masked (CP) across all MLMs.
This shows that the proposed methods are more robust against the discrepancy of word frequencies between the two groups.

\begin{figure}[t!]
    \centering
	\begin{subfigure}[b]{0.5\textwidth}
		\centering
		\includegraphics[height=1.35in]{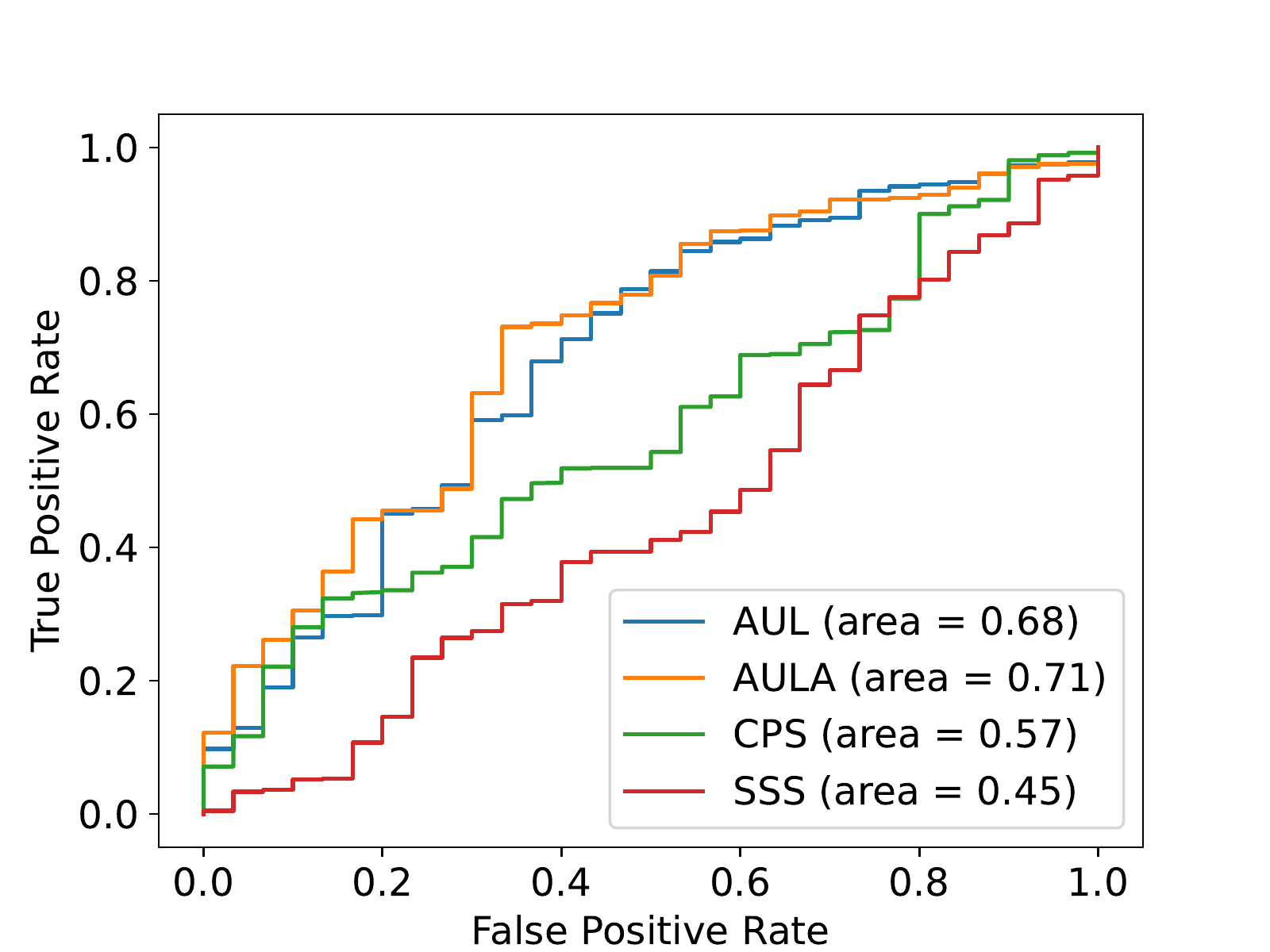}
		\caption{BERT}
		\label{fig:glove}
	\end{subfigure}
	\begin{subfigure}[b]{0.5\textwidth}
		\centering
		\includegraphics[height=1.35in]{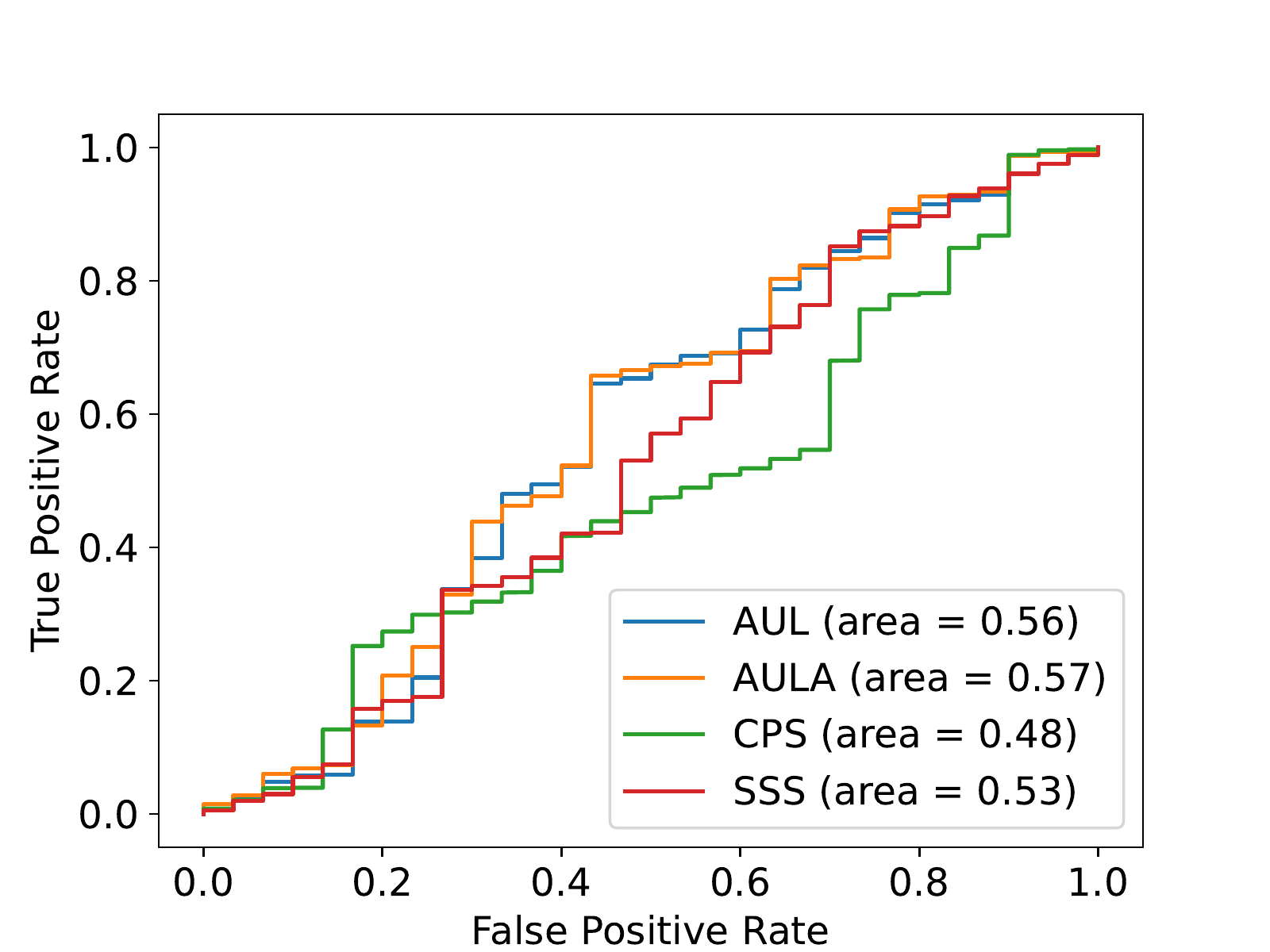}
		\caption{ALBERT}
	\end{subfigure}
\caption{ROC curve and under the curve of AUL, AULA, CPS and SSS for BERT and ALBERT on CP.}
\label{fig:roc}
\end{figure}

\begin{table*}[t]
\small
\centering
\begin{tabular}{llcc}
\toprule
Type & Example & CPS & AULA (CP) \\
\midrule
Stereotype & \textit{Immigrants} are only good at positions without a lot of human interaction. & -33.84 & \textbf{-0.01} \\
Anti-stereotype & \textit{Americans} are only good at positions without a lot of human interaction. & \textbf{-31.72} & -0.03 \\
\midrule
Type & Example & SSS & AULA (SS) \\
\midrule
Stereotype & All people from morocco are \textit{thieves} & -12.87 & -0.07 \\
Anti-stereotype & All people from morocco are \textit{kind} & \textbf{-11.41} & \textbf{-0.06} \\
Unrelated & All people from morocco are \textit{pizza} & -12.40 & -0.13 \\
\bottomrule
\end{tabular}
\caption{CPS, SSS and AULA likelihood scores for a test instance in CP (top) and SS (bottom) with BERT.}
\label{tbl:examples}
\end{table*}

\subsection{Meaningful Associations and AUL}
\label{sec:meaningful}

Recall that AUL does not mask any tokens from the test sentence.\footnote{Since AUL and AULA have the same token prediction accuracy, we report AUL only for these experiments.} 
Therefore, one might argue that the AUL might be simply filling in the masked out slot in a test sentence from the unmasked input, without considering any social biases expressed in the context.
To test whether AUL is sensitive to the meaningful associations in the input and not simply memorising the masked out tokens from test sentences, we conduct the following experiment.


On CP and SS datasets, we randomly shuffle the tokens in a test sentence and use AUL to predict the tokens as they appear in the shuffled sentence.
In \autoref{tbl:mask_noise}, we report the drop in the token-level prediction accuracy when the input is shuffled.
Because the set of tokens in a sentence is unchanged under shuffling, token frequency distribution does not affect this evaluation.
In addition, on the SS dataset, we report the drop in prediction accuracy of the unrelated candidate in each test sentence. 
From \autoref{tbl:mask_noise} we see that the token prediction accuracy drops significantly for all MLMs in both CP and SS datasets with AUL.
This result shows that AUL is sensitive to the meaningful associations in the input and not simply memorising it.



\subsection{Biases in MLMs}
\label{sec:bias-eval}


\autoref{tbl:bias_eval} shows the biases of MLMs evaluated using CPS, SSS, AUL and AULA.
All evaluation measures show that unfair social biases are learnt by the MLMs compared.
However, CPS and SSS tend to overestimate the biases compared to AUL and AULA.



\autoref{tbl:bias_cp} shows the bias scores computed using AUL (CP) and AULA (CP) have similar bias scores for each bias type in the CP dataset.
On the other hand, the bias scores computed using CPS and proposed methods have different tendencies for each bias type.
We see that the bias types with the highest and the lowest bias scores for each MLM differ between CPS and proposed methods.
For example, in BERT, the bias type with the highest CPS is \textit{sexual orientation}, whereas with proposed methods it is \textit{disability}.

Likewise, \autoref{tbl:bias_ss} shows the bias scores computed using SSS, AUL (SS) and AULA (SS) for the different types of biases in the SS dataset.
Here again we see that the bias types with the highest and the lowest bias scores for each MLM differ between SSS and proposed methods, except for \textit{gender}, which is rated as the highest scoring bias type by all measures in ALBERT.
Interestingly, SSS rated \emph{gender} and \emph{race} to be respectively the highest and lowest scoring biases in all MLMs.

We compute the agreement between the MLM-based biased scoring methods discussed in the paper and human bias ratings in CP.
Specifically, each sentence pair in CP is independently annotated by six human annotators indicating whether a particular social bias is expressed by the sentence pair.
The majority over the bias types indicated by the annotators is considered as the bias type of the sentence pair.
Considering that a sentence pair can be either biased or not (i.e.a binary outcome) according to human annotators, we model this as a binary retrieval task where we must \emph{predict} whether a given sentence pair is socially biased using an MLM-based bias scoring method.\footnote{Popular rank correlations such as Spearman/Pearson correlation coefficients are unfit for this evaluation task because human rated bias outcomes are binary, whereas MLM-based bias scores are continuous values.}
We split sentence pairs in the CP dataset into two groups depending on whether a sentence pair has received more than three biased ratings from the six annotators or not.
We then predict whether a sentence pair is biased or not at varying thresholds of an MLM-based bias score to compute the ROC\footnote{Recall that MLM bias scores are not calibrated against human ratings, hence AUC values less than 0.5 are possible.} curves shown in \autoref{fig:roc}.
Overall, we see that both AUL and AULA report higher agreement with human ratings compared to previously proposed MLM bias evaluation methods.
Moreover, CPS, which addresses the token frequency problem, does not always perform bias evaluation effectively in all MLMs compared to SSS.
\autoref{tbl:examples} shows an illustrative example where PLL scores are computed for the same input using different bias evaluation measures.
The top example shows a test instance from the CP dataset, where according to CPS there is no bias in BERT (CPS(anti-stereotype) $>$ CPS(stereotype)), whereas AULA correctly identifies this bias (AULA(anti-stereotype) $<$ AULA(stereotype)).
In fact, \textit{Americans} (116,064 occurrences) is more frequent than \textit{Immigrants} (53,054 occurrences) in our corpus.
This results in CPS reporting a high PLL score for the anti-stereotypical case, while AULA, which does not perform masking, is affected to a lesser degree, thus satisfying Criterion 2.
The bottom example shows a test instance from the SS dataset, where SSS reports a lower PLL for the stereotypical candidate than the unrelated candidate.
On the other hand, AULA correctly assigns the lowest PLL among the three candidates to the unrelated candidate, thereby satisfying Criterion 1.

\section{Conclusion}

We proposed AUL, a bias evaluation measure for MLMs using PLL where we use the \emph{unmasked} input test sentence and predict \emph{all} of its tokens. 
We showed that AUL is relatively robust against the distortions in the frequency distribution of the masked tokens,
 and can accurately predict various types of social biases in MLMs on two crowdsourced datasets.
However, AUL showed that all MLMs encode concerning social biases, and developing methods to robustly debias pretrained MLMs remains an important future research direction.
Moreover, we proposed AULA method to evaluate bias by considering tokens based on their importance in a sentence using attention weights, and showed that it matches human bias scores the most compared to other bias evaluation metrics.



\section{Ethical Considerations}

In this study, we proposed a measure for evaluating the social bias learned by MLMs.
We investigated the performance of the evaluation measures for ten different types of biases on two crowdsourced datasets.
We did not collect nor release these evaluation datasets.
The proposed evaluation measure, AUL, identified that different types of social biases are encoded in all MLMs we evaluated in the paper.
This is a worrying situation as there are already many downstream NLP applications that use these popular MLMs as-is without applying any debiasing methods.

We also note that types of biases detected by the proposed evaluation measure are limited to those annotated in the crowdsourced datasets, where annotators were limited to English speakers residing in the United States.
Therefore, the evaluations conducted in this paper do not cover cultural biases in other countries or in other languages.
This is an important limitation one must remember when interpreting the results reported in the paper.
In particular, even in the case that bias evaluation measures report an MLM to be unbiased one must be careful when deploying such MLMs in downstream applications because unfair social biases can still creep in during fine-tuning the MLMs using task-specific data.

\bibliographystyle{acl_natbib}
\bibliography{Unbiased}

\clearpage

\appendix

\section{Word frequency in Wikipedia and Books Corpus}

\autoref{tbl:word_freq} shows the frequency of eight words for each bias type in Wikipedia and BookCorpus.
Selection of these words is described in detail in \autoref{sec:adv-freq}.

\begin{wraptable}{r}{\textwidth}
\centering
 \begin{adjustbox}{width=\linewidth}
\scalebox{0.75}{
\begin{tabular}{lcccccccc}
\toprule
& \multicolumn{4}{c}{Disadvantaged group} & \multicolumn{4}{c}{Advantaged group} \\
\midrule
Race & african & jamal & asian & tyrone & american & caucasian & james & carl \\
Frequency & 278,533 & 6,292 & 148,455 & 10,382 & 1,477,133 & 7,747 & 491,860 & 73,431 \\
\midrule
Gender & she & her & women & woman & he & his & men & him \\
Frequency & 16,184,610 & 17,794,298 & 835,560 & 788,972 & 29,565,671 & 21,625,647 & 1,013,099 & 7,669,190 \\
\midrule
Sexual orientation & gay & homosexual & lesbian & bisexual & heterosexual & woman & husband & wife \\
Frequency & 82,133 & 15,121 & 23,238 & 8,500 & 6,406 & 835,560 & 310,454 & 558,139 \\
\midrule
Religion & muslim & jewish & jew & muslims & christian & christians & atheist & church \\
Frequency & 98,526 & 190,599 & 12,914 & 52,010 & 298,823 & 45,435 & 8,082 & 891,126 \\
\midrule
Age & old & elderly & teenagers & boy & young & middle & aged & adults \\
Frequency & 1,195,891 & 34,677 & 23,301 & 396,776 & 704,664 & 452,042 & 106,367 & 80,200 \\
\midrule
Nationality & mexican & italian & chinese & immigrants & american & americans & canada & canadian \\
Frequency & 107,320 & 277,740 & 293,124 & 53,054 & 1,477,133 & 116,064 & 408,676 & 295,477 \\
\midrule
Disability & ill & mentally & retarded & autistic & healthy & smart & normal & gifted \\
Frequency & 84,462 & 35,469 & 2,647 & 3,343 & 52,631 & 85,855 & 219,500 & 22,207 \\
\midrule
Physical appearance & fat & overweight & ugly & short & thin & tall & skinny & beautiful \\
Frequency & 67,664 & 6,096 & 45,210 & 637,024 & 131,631 & 171,563 & 18,457 & 302,924 \\
\midrule
Socioeconomic status & poor & ghetto & homeless & poverty & rich & wealthy & businessman & doctor \\
Frequency & 230,388 & 10,868 & 25,417 & 89,891 & 157,354 & 58,398 & 51,700 & 228,713 \\
\bottomrule
\end{tabular}
}
 \end{adjustbox}
\caption{Combined frequency of the top 8 frequent words in Wikipedia and BookCorpus.}
\label{tbl:word_freq}
\end{wraptable}

\end{document}